\documentclass{Interspeech}

\interspeechcameraready

\title{Can we train ASR systems on Code-switch without real code-switch data? Case study for Singapore's languages}

\author[affiliation={1}]{Tuan}{Nguyen}
\author[affiliation={1}]{Huy-Dat}{Tran}

\affiliation{Institute for Infocomm Research (I2R)}{A*STAR}{Singapore}
\email{nguyenvat@i2r.a-star.edu.sg, hdtran@i2r.a-star.edu.sg}

\usepackage{comment}
\usepackage{multirow}
\usepackage{multicol}
\usepackage[dvipsnames]{xcolor}
\usepackage{graphicx}  
\usepackage{array}     

\begin{document}

\maketitle

\begin{abstract}

Code-switching (CS), common in multilingual settings, presents challenges for ASR due to scarce and costly transcribed data caused by linguistic complexity. This study investigates building CS-ASR using synthetic CS data. We propose a phrase-level mixing method to generate synthetic CS data that mimics natural patterns. Utilizing monolingual augmented with synthetic phrase-mixed CS data to fine-tune large pretrained ASR models (Whisper, MMS, SeamlessM4T). This paper focuses on three under-resourced Southeast Asian language pairs: Malay-English (BM-EN), Mandarin-Malay (ZH-BM), and Tamil-English (TA-EN), establishing a new comprehensive benchmark for CS-ASR to evaluate the performance of leading ASR models. Experimental results show that the proposed training strategy enhances ASR performance on monolingual and CS tests, with BM-EN showing highest gains, then TA-EN and ZH-BM. This finding offers a cost-effective approach for CS-ASR development, benefiting research and industry.

\end{abstract}

\begin{figure*}[t!] 
    \centering
    \includegraphics[width=\textwidth]{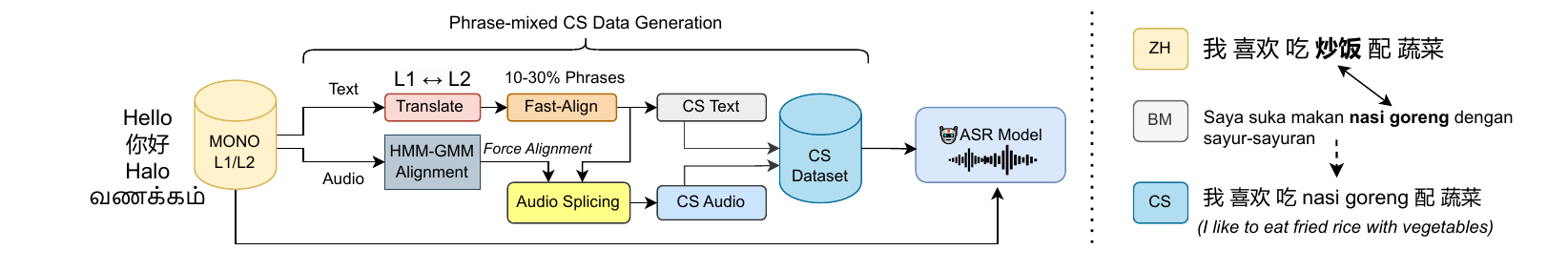} 
    \caption{Phrase-mixed CS Data Generation Pipeline for ASR Training}
    \label{fig:phrase_mix_cs_pipeline} 
    \vspace{-10pt}
\end{figure*}

\section{Introduction}

Code-switching (CS), the natural and fluid alternation between two or more languages within a single conversation or utterance, is a pervasive linguistic phenomenon worldwide, particularly as multilingualism grows \cite{newsdle_population_bilingual}. Despite its widespread nature, research on CS, especially in Automatic Speech Recognition (ASR), significantly lags behind that of monolingual language processing. Compared to the abundant resources available for monolingual corpora, CS datasets remain under-resourced \cite{shrutilipi_data}, multilingual scenarios increase complexity and lead to confusion \cite{zhang22x_interspeech}, and pose unique challenges for model architectures \cite{tok_lang_diarization, lae_st_moe, moe_cs_enc_disentanglement}. Monolingual models, trained solely on monolingual data, are demonstrably ill-equipped to handle code-switched speech \cite{one_model_rule_all}.

Current research in CS-ASR often focuses on adapting model architectures – employing techniques like dual encoders \cite{song22e_interspeech}, language-aware structures \cite{wang2024camelcrossattentionenhancedmixtureofexperts, lae_st_moe}, or language-specific attention \cite{icassp2024_int_lang_bias} – primarily to mitigate language confusion. These methods frequently necessitate retraining entire networks with limited CS data, leading to increased complexity and limited scalability. Another research direction explores cross-lingual transfer and unsupervised or self-supervised learning (SSL) leveraging large-scale pretrained datasets \cite{icassp2024_data_collect_and_arabic_asr, whisper_seame_asru_cs, indo_eng_cs_ssl}. While promising, these works often remain focused on linguistic exploration and small-scale datasets. Indeed, deep investigations into CS-ASR are largely confined to a few well-resourced language pairs, most notably Mandarin-English \cite{tok_lang_diarization, wang2024camelcrossattentionenhancedmixtureofexperts}. Consequently, much of the existing CS-ASR research leans towards exploratory studies rather than solutions that are ready for robust, industrial applications.

This paper targets the development of CS-ASR technologies for industrial use, focusing on large-scale training methodologies, particularly for under-resourced Southeast Asian (SEA) languages. We address the critical bottleneck in CS-ASR: the limited availability of large-scale, transcribed code-switching data. This scarcity is inherent due to factors such as the early stage of CS corpus development, domain-specific challenges, language pair variations, community biases \cite{jsalt_2022_cs_report}, and the inherent difficulties in data collection and annotation \cite{chi23_interspeech}.

To address limited data, we propose using large synthetic datasets. Our method introduces a phrase-level mixing technique (improving traditional lexicon-based approaches \cite{speech_collage}) to create natural-sounding code-switched speech by blending phrases from monolingual sources \cite{language_mixing_singaporean}. We built test sets by leveraging ChatGPT-generated conversations (text) spoken by Singaporeans for BM-EN and ZH-BM (within-sentence mixing), and a synthetic sentence-mixed TA-EN test set. Testing three leading large-scale pretrained ASR models on these sets revealed their performance in real-world multilingual industrial scenarios. This work is the first comprehensive benchmark of SOTA ASR models for under-resourced SEA languages, providing scalable solutions for research and industry.

Summarizing our experiments, demonstrate the effectiveness of our phrase-mixed data augmentation and benchmarks. For BM-EN, our findings reveal remarkable distributional alignment with real-world test sets, directly resulting in strong, validated real-world performance and powerfully confirming our synthetic data's fidelity in bridging the domain gap. While ZH-BM and TA-EN pairs show encouraging alignment suggesting a generalizable approach, future progress lies in integrating language-specific linguistic characteristics into phrase-mixing to optimize distributional matching and real-world performance for diverse pairs. Our newly established benchmark on under-resourced language pairs, including BM-EN, ZH-BM, and TA-EN, underscores the efficacy of large-scale pre-training on state-of-the-art models, with fine-tuned SeamlessM4T-v2-Large outperforming competitors (Whisper-Large-v3-Turbo and MMS-1B-All) and emerging as a leading candidate for industrial-scale CS-ASR. 

\noindent In summary, the main contributions of this paper are:

\begin{itemize}
    \item \textbf{Scalable Data Augmentation:} We introduce a novel and scalable phrase-mixed data augmentation method, demonstrating significant performance gains on real-world code-switching ASR test sets.
    \item \textbf{Comprehensive SOTA Model Analysis:} We present a detailed comparative analysis of three leading state-of-the-art ASR models (\textsc{Whisper-Large-Turbo-V3}, \textsc{MMS-1B-All}, and SeamlessM4T-v2-Large), revealing consistent performance improvements across multiple under-resourced language pairs.
    \item \textbf{Publicly Available Benchmark Datasets:} We release novel evaluation test sets for three under-resourced language pairs to foster research in code-switching ASR. Datasets and model APIs are available upon email request.
\end{itemize}

\section{Methods}

\subsection{Code-Switching in Singapore}

Singapore is a culturally and linguistically diverse nation, representing the broader Southeast Asian region within a single country. The primary language of communication is English, which is supported by government policies that promote Mandarin Chinese, Bahasa Malay, and Tamil as official languages \cite{sing_langs}. As a result, a significant portion of the population is bilingual or multilingual. In both casual and formal contexts, it is common for individuals to code-switch, interweaving two or more languages within a single sentence to facilitate communication and express cultural identity \cite{bilingual_policy_nlb}.

This study focuses on the languages spoken in Singapore, specifically Singlish (EN), Mandarin Chinese (ZH), Bahasa Malay (BM), and Tamil (TA). We conduct experiments on under-researched code-switching language pairs, including BM-EN, ZH-BM, and TA-EN. We exclude EN-ZH, as it's already well-studied.

\subsection{Phrase-Mixed: Enhancing Naturalness in Scalable Synthetic Code-Switch Data Generation}

Building upon Speech Collage \cite{speech_collage}'s approach of leveraging monolingual corpora for synthetic speech creation, we introduce several enhancements to improve synthesis quality (detail of our pipeline in Figure \ref{fig:phrase_mix_cs_pipeline}):

\begin{enumerate}

    \item \textbf{Translation:} We employ Google Translate for BM-EN and TA-EN pairs, and Mesolitica's translation model \footnote{hf.co/mesolitica/translation-nanot5-base-malaysian-cased} for ZH-EN, to optimize for speed and accuracy.
    
    \item \textbf{Textual Alignment:} We replace BERT-based alignment with \textsc{Fast\_Align} \cite{fast_align_text} for improved scalability.
    
    \item \textbf{Phrase-Mixed Replacement:} Singaporean code-switching occurs primarily at the phrase level rather than the word level \cite{language_mixing_singaporean}, with speakers naturally alternating between full phrases while maintaining academic terminology. To better reflect this phenomenon, we replace the fixed 20\% lexicon substitution with a more flexible 10-30\% mixing per sentence, capturing more authentic conversational patterns that incorporate both phrases and individual words.
    
    \item \textbf{Audio Splicing:} We improve Speech Collage \cite{speech_collage} by incorporating NeMo's amplitude-based normalization \footnote{\url{github.com/NVIDIA/NeMo/tree/main/scripts/speech\_recognition/code\_switching}}, which produces more natural speech output. This adjustment addresses the distortion observed when using the original energy-based normalization.
    
\end{enumerate}

\subsection{ASR Systems Training}

\textbf{Large-Scale Pre-Trained Models.} Recent advances in large-scale pre-trained speech models enable adaptation to downstream tasks with limited data. Whisper \cite{whisper_model}, MMS \cite{mms1b_model}, and SeamlessM4T \cite{seamless_model} achieve state-of-the-art performance across diverse speech tasks and low-resource settings, benefiting from advanced architectures and extensive training data \cite{ml_superb2}.

Whisper \cite{whisper_model}, a fully supervised model, supports approximately 100 languages and was initially trained on 680k hours of data, later extended to 5 million hours of weakly labeled data in version 3. MMS \cite{mms1b_model} extends wav2vec2 to 1,000 languages via self-supervised pretraining and language-specific adaptation. SeamlessM4T \cite{seamless_model}, based on w2v-BERT, is trained on 4.5 million hours and integrates speech-text alignment. These models serve as strong baselines for ASR research, allowing us to emulate industrial-scale systems.

\textbf{Fine-Tuning and Adaptation.} We fine-tune Whisper and SeamlessM4T on multilingual tasks (EN, ZH, BM, TA) and code-switching pairs (BM-EN, ZH-BM, TA-EN). For MMS, we explore adapter-based and full-model fine-tuning with vocabulary adjustments. Our results indicate that training on multilingual data with code-switching outperforms purely multilingual training.


\begin{table*}
    \centering
    \caption{WER/CER/MER for different models on monolingual and code-switch datasets. Relative improvement indicate in round brackets, either improve (green) or decrease (red) over baseline.}
    \resizebox{\textwidth}{!}{%
    \begin{tabular}{c l c|cccc|lll|c|c}
    \toprule
    \textbf{Data} & \textbf{Model} & \textbf{\#} & \multicolumn{4}{c|}{\textbf{Multilingual}} & \multicolumn{3}{c|}{\textbf{Code-Switch}} & \textbf{MONO} & \textbf{CS} \\
    &  &  & \textbf{EN} & \textbf{ZH} & \textbf{BM} & \textbf{TA} & \textbf{BM-EN} & \textbf{ZH-BM} & \textbf{TA-EN} &  &  \\ \midrule
    \textsc{Baseline}  & \textsc{WhisperTurbo-v3}  & 0.8B  & \textbf{22.04}   & 21.55   & 39.17 & 58.31 & 30.67 & 58.67 & 114.6 & 35.27   & 67.97    \\
    + \textsc{mono}  & | \textsc{FT} & 0.8B      & 28.41   & 13.48   & 17.44 & 37.56  & 30.45 \textcolor{Green}{(0.7)} & 76.14 \textcolor{Red}{(22.9)}  & 188.9 \textcolor{Red}{(39.3)} & 24.22   & 98.50   \\
    \quad + \textsc{cs} & | \textsc{FT}  & 0.8B      & 28.97   & \textbf{12.85}   & \textbf{16.62} & \textbf{23.40}  & \textbf{16.75} \textcolor{Green}{\textbf{(83.1)}} & \textbf{58.22} \textcolor{Green}{\textbf{(0.8)}} & \textbf{51.37} \textcolor{Green}{\textbf{(123)}} & \textbf{20.46} & \textbf{42.11}   \\ \midrule
    \textsc{Baseline} & \textsc{MMS-1B-All}  & 0.97B  & 25.10    & 39.52   & 49.03 & 42.72   & 48.62 & 67.31 & 72.26 &  39.09    & 62.73  \\
    + \textsc{mono} & | Adapter & 0.97B  & \textbf{24.07}  & 31.92   & 42.09 & 45.53  & 48.44 \textcolor{Green}{(0.4)} & 65.21 \textcolor{Green}{(3.2)} & 73.65 \textcolor{Red}{(1.9)} & 35.90  & 62.43  \\
    \quad + \textsc{cs} & | Adapter & 0.97B  & \textbf{24.07}  & 31.11   & 41.84 & 45.31  & 44.61 \textcolor{Green}{(9)} & 56.03 \textcolor{Green}{(20.1)} & 73.98 \textcolor{Red}{(2.3)} & 35.58   & 58.21  \\
    \quad + \textsc{cs}  & | \textsc{FT} w/ Char & 0.97B & 32.06 & \textbf{20.50} & 26.54 & 35.42 & 38.05 \textcolor{Green}{(27.8)} & 50.95 \textcolor{Green}{(32.1)} & 41.23 \textcolor{Green}{(75.3)} & 28.63 & 43.41  \\
    \quad + \textsc{cs} & | \textsc{FT} w/ BPE & 1.03B & 24.45 & 21.05 & \textbf{23.56} & \textbf{34.98} & \textbf{23.18} \textcolor{Green}{\textbf{(109.7)}} & \textbf{43.26} \textcolor{Green}{\textbf{(55.6)}} & \textbf{35.40} \textcolor{Green}{\textbf{(104.1)}} & \textbf{26.01} & \textbf{33.95}   \\ \midrule
    \textsc{Baseline} & \textsc{SeamlessM4T-v2} & 1.5B   & 28.07   & 25.38   & 33.39 & 30.41  & 38.36 & 57.06 & 63.42  & 29.31 & 52.95  \\
    + \textsc{mono} & | \textsc{FT} &  1.5B      & 12.41   & 9.52    & 15.83 & 21.78  & \textbf{13.08} \textcolor{Green}{\textbf{(193.3)}} & 71.10  \textcolor{Red}{(19.7)} & 58.80 \textcolor{Green}{(7.9)} & 14.89 & 47.66 \\
    \quad + \textsc{cs}  & | \textsc{FT}  & 1.5B  & \textbf{12.30} & \textbf{9.45} & \textbf{15.67} & \textbf{21.51} & 14.58 \textcolor{Green}{(163.1)} & \textbf{39.74} \textcolor{Green}{\textbf{(43.6)}} & \textbf{38.58} \textcolor{Green}{\textbf{(64.4)}} & \textbf{14.73} & \textbf{30.97} \\
    \bottomrule
    \end{tabular}%
    }
    \label{tab:experimental_results}
    \vspace{-10pt}
\end{table*}

\section{Experiments}

\subsection{\textsc{Mono} Data}

We utilize monolingual data from four languages: EN, ZH, BM, and TA, incorporating both local and non-local accents. Local accents refer to those associated with Singapore or similar regions. All datasets, except for the \textbf{Internal} dataset (which is sampled from our proprietary data), are listed in Table \ref{tab:dataset_info}. We merged all the data, and refer to this dataset as \textsc{mono}.

\begin{table}
\centering
\caption{MONO Dataset Summary by Language}
\renewcommand{\arraystretch}{1.3}  
\resizebox{\columnwidth}{!}{%
\begin{tabular}{llr}
\hline
\textbf{Dataset} & \textbf{Lang} & \textbf{Hrs} \\
\hline
\textbf{LibriSpeech} \cite{librispeech_data} (250h) + \textbf{IMDA} \cite{imda_dataset}\{1-3, 5-6\} (750h) & EN & 1000 \\ \hline
\textbf{AIShell} \cite{aishell_dataset} (180h) + \textbf{MagicData} \cite{magic_data} (360h) & ZH & 1000 \\
+ \textbf{WenetSpeech} \cite{wenetspeech_data} (180h) + \textbf{Internal} (280h) & & \\ \hline
\textbf{IISc-MILE} \cite{iisc_mile_data} (77h) + \textbf{CV19} \cite{cv_dataset} (82h) & TA & 775 \\
+ \textbf{Shrutilipi} \cite{shrutilipi_data} (410h) + \textbf{Internal} (206h) & & \\ \hline
\textbf{Internal} & BM & 1000 \\
\hline
\textbf{Total} & & \textbf{3775} \\
\hline
\end{tabular}
}
\label{tab:dataset_info}
\end{table}

\subsection{Phrase-mixed Code-Switch data generation}

\begin{table}[h]
\centering
\caption{\textsc{CS}-related statistical information of our training phrase-mixed, our test sets, and real \textsc{SEAME} testset.}
\resizebox{\columnwidth}{!}{%
\begin{tabular}{lrrrr}
\toprule
& Hours & CMI & I-Index & M-Index \\
\midrule
\multicolumn{5}{l}{\textit{Phrase-mixed Train sets - \textsc{CS}}} \\
BM-EN Phrase-Mixed & 1000 & 38.37 & 46.81 & 23.47 \\
ZH-BM Phrase-Mixed & 1000 & 14.87 & 19.03 & 14.68 \\
TA-EN Phrase-Mixed & 1000 & 31.45 & 37.33 & 22.64 \\
\midrule
\multicolumn{5}{l}{\textit{Our test sets}} \\
Real BM-EN & 11.61 & 36.88 & 39.44 & 24.73 \\
Real ZH-BM & 7.46 & 18.18 & 16.80 & 21.40 \\
Sentence-Mixed TA-EN & 13.73 & 19.93 & 6.50 & 22.63 \\
~~~~\textit{MUCS (TA) \cite{mucs_ta_dataset} + IMDA3 (EN) \cite{imda_dataset}} & & & & \\
\midrule
\multicolumn{5}{l}{\textit{Real CS Testset}} \\
SEAME EN-ZH dev-man/-sge \cite{seame_data} & 11.40 & 15.40 & 14.17 & 14.62 \\
\bottomrule
\end{tabular}
}
\vspace{-15pt}
\label{tab:cs_stats}
\end{table}

Using the \textsc{mono} dataset, we generate code-switching text through translation (L1 → L2) and alignment (via \textsc{Fast\_Align}), as detailed in Section 2.2. We then replace 10–30\% of tokens per sentence under a consecutive word constraint to mimic natural phrase-level switching. For audio generation, we align words to speech by training a simple HMM-DNN model \cite{hmm_alignment}, then splice segments guided by code-switched text, applying amplitude normalization for naturalness.

We generate 1,000 hours per language pair (BM-EN, ZH-BM, TA-EN), totaling 3,000 hours of synthetic code-switched data (called \textsc{CS}). Note that we do not know the real characteristics for each pair in advance.

\subsection{Evaluation Settings}

\noindent\textbf{Multilingual Evaluation.} We conducted comprehensive evaluations by averaging results across language-specific test sets containing general and local accents (except Tamil, lacking local accent data). test sets include:
\begin{itemize}
\item \textbf{English}: LibriSpeech test-other (noisy reading) \cite{librispeech_data}, and internal datasets: clean scripts and noisy conversations.
\item \textbf{Chinese}: AiShell-test (reading) \cite{aishell_dataset}, MagicHub (conversation) \cite{magic_data}, and an internal test set in diverse conditions.
\item \textbf{Malay}: Low-resource internal sets (noisy, conversational, scripted).
\item \textbf{Tamil}: Mile Tamil (clean reading) \cite{iisc_mile_data}, OpenSLR 65 (clean reading) \cite{openslr_ta_65}, and internal reading scripts.
\end{itemize}

\noindent\textbf{Code-Switch Evaluation.} We evaluated code-switching models using real-world test sets (BM-EN, ZH-BM) collected from ChatGPT conversations (text) spoken by Singaporeans. Due to resource constraints, Singaporean-spoken transcribed data for TA-EN was unavailable. Instead, we synthesized TA-EN test set by inter-sentential switching sentences from the MUCS (TA) \cite{mucs_ta_dataset} and IMDA3 (EN conversational) \cite{imda_dataset} datasets using NeMo's sentence-level CS scripts. These test sets are the first benchmarks for code-switch evaluation in under-resourced Southeast Asian languages, offering valuable insights and advancing research in this area.

We use WER for all languages and pairs except ZH and ZH-BM, where we use CER and Mixed Error Rate (MER) due to their character-based writing system. For fair comparison across models, we split test audio longer than 30s using PyannoteVAD \cite{pyannote_vad} to accommodate Whisper's context limit \cite{whisper_model}.

\subsection{Finetuning Setup}


\textbf{Whisper \& SeamlessM4T.} Both autoregressive models use BPE tokenizers and were fine-tuned. For Whisper, we used Whisper-Large-v3-Turbo (4-layer decoder – \textsc{WhisperTurbo-v3}) with its language prefix system, phrase-mixed samples (\textsc{SC}) were assigned prefixes by priority: \textbf{BM}-EN, \textbf{TA}-EN, \textbf{ZH}-BM. For SeamlessM4T, we use SeamlessM4T-v2-large (\textsc{SeamlessM4T-v2}), the largest version available.

\textbf{MMS.} An encoder-only model trained with CTC loss and character outputs, we used \textsc{MMS-1B-All} (fine-tuned on over 1,000 languages). We experimented with both adapter-based fine-tuning (each language's adapter) and full fine-tuning (averaging adapter weights across four languages), using either a merged character-level vocabulary or a BPE vocabulary (via the Whisper tokenizer).

All models were trained for 3 epochs with a learning rate of 1e-5, 20\% linear warm-up, cosine decay, and the AdamW optimizer (weight decay = 0). We applied 20\% speed perturbation and 20\% MUSAN noise ([10, 30]dB). Note that we did not use SpecAugment for SeamlessM4T due to minimal gains. All experiments utilized the Transformers library.

\section{Results}

\begin{figure*}[t] 
    \centering
    \includegraphics[width=\textwidth]{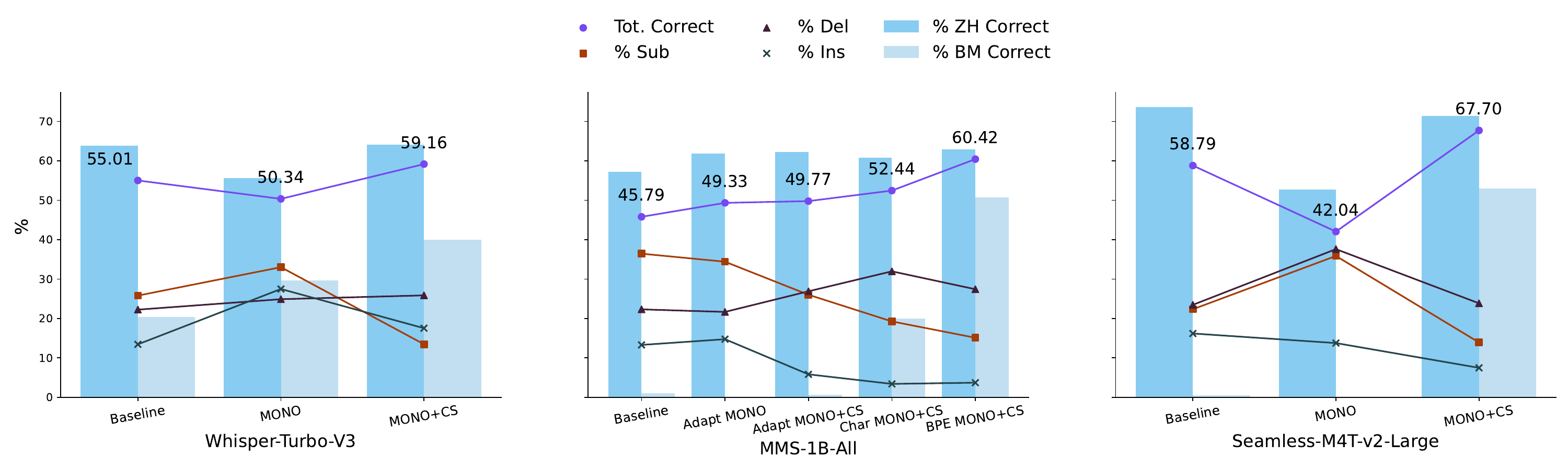} 
    \caption{MER components of ZH-BM testset. Include Correctness, Insertion rate, Deletion rate, and Substitution rate.}
    \label{fig:vendor_zh_bm_img} 
\end{figure*}

\subsection{Analysis on Code-Switch Characteristics in Phrase-mixed Training Sets and Test Sets}
\label{subsec:analyst_cs_stats}

We assessed code-switching (CS) patterns using three metrics: the Code-Mixing Index (CMI) considers both the number of switches and language distribution \cite{speech_collage}; the I-Index for how often switching occurs; and the M-Index for language balance \cite{jsalt_2022_cs_report}. Table~\ref{tab:cs_stats} shows these metrics for each dataset, averaged across utterances.  The BM-EN training and test sets are very similar across all metrics (CMI: 38.37 vs. 36.88, I-Index: 46.81 vs. 39.44, M-Index: 23.47 vs. 24.73).  However, for ZH-BM, the training and test sets are less consistent. The training set shows higher switching frequency but lesser language balance compared to the test set. In contrast, compared to training data, the sentence-mixed TA-EN test set exhibits a similar language balance (M-Index $\approx 22.63$) but a lower switching frequency (I-Index: 6.50 vs. 37.33) due to its inter-switched nature. The SEAME EN-ZH benchmark \cite{seame_data} has a lower code-switching complexity (CMI: 15.40), suggesting recognition tasks might be less challenging \cite{cmi_compare_egyptian}. These scores suggest that our phrase-mixing method somewhat captures real code-switching patterns, which is beneficial for training models, especially when real code-switched examples are scarce. Test sets like BM-EN and ZH-BM seem challenging enough to represent real-world situations. However, TA-EN, with its sentence-level switching, may be less challenging to process.

\subsection{Multilingual ASR}

Table \ref{tab:experimental_results} shows that fine-tuning generally improves performance across languages for both \textsc{MONO} and \textsc{MONO+CS} setups. Exceptions include \textsc{WhisperTurbo-v3} (\textsc{MONO} and \textsc{MONO+CS})—which degrades in English, particularly on the noisy dataset with local accents—and the \textsc{MMS-1B Character} model, which underperforms on both NLB and LibriSpeech test-other (U.S. accents). For non-English languages, fine-tuning with \textsc{MONO} enhances performance, with further gains when incorporating \textsc{CS}, indicating that more fine-tuning data is beneficial. Notably, \textsc{SeamlessM4T-v2} shows the largest improvements relative to \textsc{WhisperTurbo-v3} and \textsc{MMS-1B-All}. 

\subsection{Code-Switch ASR on BM-EN, ZH-BM, TA-EN}

Table~\ref{tab:experimental_results} demonstrates consistent performance improvements across all models in the \textsc{MONO+CS} setting, validating the efficacy of augmenting monolingual data with phrase-mixed code-switched corpora. Relative gains vary by language pair: BM-EN shows the largest improvement, followed by TA-EN and ZH-BM, aligning with the linguistic analysis in Section~\ref{subsec:analyst_cs_stats}: BM-EN had strong alignment on train-test, TA-EN show both language-balance and less switching frequency, and ZH-BM have the big gap on language balance, make the model harder to predict.

For BM-EN, \textsc{MONO+CS} enhances all models except \textsc{SeamlessM4T-V2}, which slightly degrades despite gains in \textsc{MONO}. TA-EN exhibits different challenges: \textsc{WhisperTurbo-v3} suffers \textgreater100\% WER in \textsc{Baseline} and \textsc{MONO} due to high insertion errors, which is solved by \textsc{MONO+CS}. \textsc{MMS-1B-All} requires full finetuning (\textsc{BPE-based} outperforms \textsc{Char-based}) for improvements, showing no gains with adapter-based approaches.

ZH-BM yields minimal \textsc{MONO+CS} gains. While \textsc{WhisperTurbo-v3} and \textsc{SeamlessM4T-v2} degrade in \textsc{MONO}, \textsc{MMS-1B-All} remains stable. Figure~\ref{fig:vendor_zh_bm_img} reveals model-specific biases: \textsc{WhisperTurbo-v3} marginally improves BM recognition, but is affected by high insertion/deletion errors; \textsc{MMS-1B-All} and \textsc{SeamlessM4T-v2} initially fail to detect BM (suggest a strong bias in ZH) in \textsc{MONO} but show substantial error reduction (I/D/S) after \textsc{MONO+CS} fine-tuning.

These results underscore the necessity of code-switched data to mitigate monolingual biases and improve robustness in mixed-language scenarios.

\section{Conclusion}

This study demonstrates that high-performing ASR systems for multilingual and code-switched speech can be effectively trained without real code-switching data. By leveraging phrase-mixed synthetic data validated on new benchmarks for Singaporean language pairs (BM-EN, ZH-BM, TA-EN) to successfully model real-world code-switching patterns. Key results showed that for BM-EN, strong alignment between synthetic training and real test data drove the largest relative improvements of approximately 83\% with \textsc{WhisperTurbo-v3}, while for ZH-BM, the approach mitigated Mandarin bias in pretrained models, achieving around 43.6\% improvement with \textsc{SeamlessM4T-v2}. Furthermore, significant gains were achieved on TA-EN despite resource constraints, highlighting the method's adaptability. This scalable and cost-effective approach is particularly beneficial for under-resourced languages and multilingual culture like Singapore, where collecting annotated code-switched data is challenging. Future work should focus on enhancing the linguistic realism of synthetic data, for example by integrating prosody or syntactic rules, and expanding benchmarks to include underrepresented SEA languages.

\bibliographystyle{IEEEtran}
\bibliography{mybib}

\end{document}